\documentclass[10pt,a4paper,twocolumn]{article}

\usepackage[utf8]{inputenc}
\usepackage[T1]{fontenc}
\usepackage{lmodern}
\usepackage[left=16mm,right=16mm,top=20mm,bottom=20mm,columnsep=6mm]{geometry}
\usepackage{graphicx}
\usepackage[authoryear,round]{natbib}
\usepackage{amsmath,amssymb}
\usepackage{algorithm}
\usepackage{algpseudocode}
\usepackage{booktabs}
\usepackage{xurl}
\usepackage{microtype}
\usepackage{titlesec}
\usepackage[unicode,hidelinks]{hyperref}

\titleformat{\section}{\large\bfseries\raggedright}{\thesection}{0.6em}{}
\titleformat{\subsection}{\normalsize\bfseries\raggedright}{\thesubsection}{0.6em}{}
\titleformat{\subsubsection}{\normalsize\itshape\raggedright}{\thesubsubsection}{0.6em}{}
\hypersetup{
  pdftitle={Reconstruction-Based Diagnosis through G\"odel-Regularized Virtual Observability},
  pdfauthor={Oliver Niggemann, Jonas Ehrhardt, Alexander Diedrich}
}

\title{Reconstruction-Based Diagnosis through\\G\"odel-Regularized Virtual Observability}

\author{Oliver Niggemann\quad Jonas Ehrhardt\quad Alexander Diedrich\\[0.4em]
\small Chair of Computer Science in Mechanical Engineering,\\
\small Helmut Schmidt University, Hamburg, Germany\\
\small (e-mail: oliver.niggemann@hsu.hamburg).}
\date{}

\begin{document}
\twocolumn[{
\begin{@twocolumnfalse}
\maketitle

\begin{abstract}
Not every component in a technical system can be equipped with sensors. 
Hence, when a fault occurs, available sensors often detect it only after its effects have propagated. 
This makes the isolation of faults difficult.
We approach this problem by introducing virtual sensors. 
For each combination of component and fault type, we let external knowledge describe the fault-sensitive variables.
We introduce a novel neural observer that is only trained on normal data. 
For fault isolation, it uses available sensor measurements where it is possible and reconstructs missing variables as virtual sensors to predict normal behavior. 
It detects deviations between current values and nominal predictions and computes local symptoms, which are then compared by onset time.
The result is a set of root causes describing the faulty behavior. 
We use G\"odel logic to encode physical knowledge and guide training.
We evaluate our approach on synthetic networks, the Tennessee Eastman Process, and datasets of two modular production plants.
\end{abstract}

\begin{center}
\end{center}
\vspace{1em}
\end{@twocolumnfalse}
}]

\section{Introduction}
\label{sec:introduction}

Fault diagnosis supports the safe and reliable operation of technical processes. 
But not all behaviors within a system can be observed at all times. 
A fault may hence appear at an unobserved location and its effects then propagate through material, energy, and control connections before a sensor can detect them. 
Noise and delays can change the order of symptoms. 
Model-based methods address this with residual and structural models \citep{isermann2005model,krysander2008efficient}, while data-driven methods learn normal or faulty behavior from observations \citep{qin2012survey}. Both must infer the cause from indirect and delayed evidence.


Suppose that every possible root cause had a fault-sensitive local sensor that reacts before all propagated symptoms. 
The first persistent local symptom would then identify the root cause, and fault isolation would reduce to comparing symptom onset times. 
However, no real-world plants offer such complete coverage, and learned estimates remain uncertain. 
In this article we address how machine learning can close the observation gaps without hiding this uncertainty.

We propose Reconstruction-Based Diagnosis through Virtual Observability (ReDiVO) 
which employs virtual sensors to make the whole system observable.
We use available a-priori knowledge to determine the components within the system and associate these with known sensor locations. 
We then identify locations that contain components but have no direct sensors monitoring the specific components.  
For instance, if some pump performance is not measured, we reconstruct it from related process measurements like downstream flow and control signals.
Measured and virtual sensors together create \emph{virtual observability}, this means that every fault candidate has at least one observed or virtual root-cause variable. 
We propose a neural observer which compares the current value of each variable with a nominal prediction learned from normal data, both as uncertainty ranges. 
A persistent separation of these ranges is a local symptom. 
We compare the onset times of these symptoms and return a set of candidate diagnoses.

We pursue a hybrid approach fusing data-driven modeling of normal behavior with a-priori knowledge. 
Without reference values, normal data alone may not determine a reconstruction uniquely. 
A shifted or rescaled estimate of the pump performance, for example, can predict the measured flows equally well. 
We therefore add physical knowledge about topology, qualitative dependencies, balances, and operating constraints, expressed as rules in G\"odel logic. During training, rule violations penalize the reconstructions and nominal predictions. 
For example, a higher pump performance should not reduce downstream flow when all other influences stay fixed. 
Such a rule favors plausible estimates without forcing the pump to stay nominal.

We address three research questions. 
\textbf{RQ1}: Can a virtual sensor reconstruct an unobserved component variable well enough to provide a useful local symptom? 
\textbf{RQ2}: Which types of prior knowledge in G\"odel logic improve this reconstruction? 
\textbf{RQ3}: When does the first local symptom on a common clock isolate the root cause?

We make four contributions. 
First, we formulate diagnosis under incomplete observability as virtual observability of component-fault candidates. 
Second, we develop a neural observer that learns reconstruction and open-loop prediction from normal data only. 
Third, we introduce a training-only regularizer in G\"odel logic that encodes topology, qualitative relations, and physical constraints without excluding the fault effects of interest. 
Fourth, we turn calibrated current and nominal value ranges into persistent symptoms and use their onset intervals to return a unique candidate, an ambiguity group, or no diagnosis. 
We evaluate ReDiVO on synthetic component networks, the Extended Tennessee Eastman Process~\citep{bathelt2015revision,reinartz2021extended}, HAI-CPPS, and TwinFlow~\citep{hranisavljevic2026twinflow}.

\section{Related Work}
\label{sec:related}

Structural diagnosis isolates faults through overdetermined model relations \citep{krysander2008efficient}, and diagnosability determines whether observations distinguish a fault within bounded time \citep{sampath1995diagnosability}. 
Set-based methods compare measurements with all values allowed by uncertain models \citep{harirchi2018guaranteed}. 
We adopt this range-based view for learned virtual sensors and symptom onset times, but use empirical bounds from normal data that do not need to hold under faults. 
On the data-driven side, subsystem-wise anomaly detectors can feed consistency-based diagnosis \citep{steude2024diagnosis}, and causal methods reason over monitored variables with anomaly times or Granger causality \citep{assaad2023easyrca,han2025aerca}. We address the complementary case in which a fault-relevant variable is not monitored and must first be reconstructed as a virtual sensor.

Learned observers estimate unmeasured variables from measured histories \citep{revach2022kalmannet}, and trained networks can reconstruct missing inputs by optimization \citep{ludwig2025virtualsensors}. SMT reasoning over critical pairs infers measurements that make faults distinguishable, but requires an SMT system model \citep{diedrich2024sensor}. These methods neither assign variables to component-fault candidates nor compare symptom onset times. To guide learning, differentiable fuzzy logic turns symbolic knowledge into training losses \citep{badreddine2022logic,vanKrieken2022analyzing}, also for fault diagnosis \citep{hogea2024logiclstm}, and system descriptions for diagnosis can be learned from data \citep{diedrich2024learning,moddemann2024extracting}. We combine both research directions: a-priori system knowledge constrains reconstructions and nominal predictions during training but is not used as online evidence.
\section{Reconstruction-Based Diagnosis through Virtual Observability}
\label{sec:method}
 
\subsection{Diagnosis Problem and Required Conditions}
\label{sec:diagnosis-problem}
 
Let $\mathcal C$ be the finite set of physical components. For each component $c\in\mathcal C$, the finite set $\mathcal A_c$ contains the considered fault types. A diagnosis candidate $r=\langle c,a\rangle$ is a hypothesis about the root cause of a fault: component $c$ has fault type $a$. The candidate set $\mathcal R$ contains all hypotheses that diagnosis distinguishes:
\begin{equation}
\mathcal R=\{\langle c,a\rangle\mid c\in\mathcal C,
a\in\mathcal A_c\}.
\label{eq:diagnosis-problem}
\end{equation}
Each process variable has an index $j$, and $v_j(t)$ denotes its value at sample $t$. A variable is \emph{observed} if a sensor measures it online and \emph{unobserved} otherwise. Before training, domain experts or plant documentation provide \emph{a-priori system knowledge} about components, fault-sensitive variables, and reconstruction dependencies. For each diagnosis candidate $r\in\mathcal R$, this knowledge specifies a nonempty set $B_r$ of fault-sensitive \emph{root-cause variables}. Each of them is observed or reconstructed as a \emph{virtual sensor}, that is, a learned estimate of an unobserved variable. A \emph{local symptom} of $r$ is a persistent deviation in a variable of $B_r$. ReDiVO achieves \emph{virtual observability} if every candidate has at least one observed or reconstructed root-cause variable.
 
Learning uses only normal time series recorded during fault-free operation. A-priori knowledge identifies the predictor variables used as inputs for reconstruction. Offline reference values, where available, supervise reconstruction during training but are not online inputs. For example, a temporary sensor may record the pump speed during training-data collection even if that measurement is unavailable during deployment. Online, ReDiVO receives the measurements and those commands or setpoints that do not reveal the fault. It returns either one candidate, an ambiguity group (Section~\ref{sec:diagnosis}), or no diagnosis.
 
 
\subsection{Learning Virtual Observability}
\label{sec:observer}
 
\begin{figure*}[t]
\centering
\includegraphics[width=\textwidth]{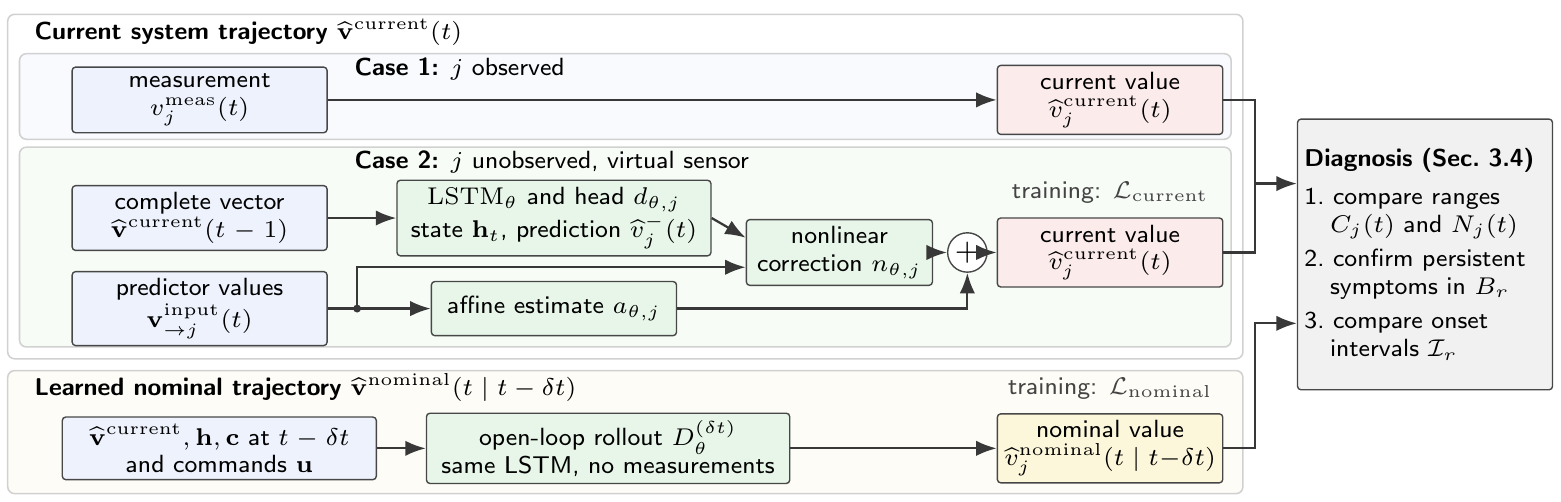}
\caption{Neural observer and diagnosis. The current trajectory uses measurements where available and virtual sensors otherwise. The nominal trajectory rolls the same network forward from $t-\delta t$ without measurements. Both losses apply only during training on normal data.}
\label{fig:redivo}
\end{figure*}

ReDiVO compares two estimates of all process variables (Fig.~\ref{fig:redivo}). The \emph{current system trajectory} $\widehat{\mathbf v}^{\mathrm{current}}(t)$ estimates what the plant does at sample $t$. It takes the measurement of observed variables (Case~1) and reconstructs unobserved variables (Case~2). The \emph{learned nominal trajectory} $\widehat{\mathbf v}^{\mathrm{nominal}}(t\mid t-\delta t)$ estimates what the plant does under normal behavior. It starts $\delta t$ samples earlier and runs a long-short-term memory (LSTM) network on its own predictions, without new measurements. A fault can separate both trajectories (Section~\ref{sec:diagnosis}). 
 
For the \emph{current system trajectory}, each unobserved variable $j$ is reconstructed from \emph{predictors} identified by the a-priori system knowledge (Case~2 in Fig.~\ref{fig:redivo}). Predictors may include internal process quantities. Their values form the input vector $\mathbf v_{\rightarrow j}^{\mathrm{input}}(t)$. An observed predictor $k$ contributes its measurement, and an unobserved one its one-step prediction $\widehat v_k^{-}(t)$, which uses information only up to $t-1$. The target $j$ is never its own predictor. Both choices prevent circular reconstruction. The current system trajectory is
\begin{equation}
\begin{aligned}
(\mathbf h_t,\mathbf c_t)
 &=\operatorname{LSTM}_{\theta}
 (\widehat{\mathbf v}^{\mathrm{current}}(t-1),
 \mathbf h_{t-1},\mathbf c_{t-1}),\\
\widehat v_j^{-}(t)&=d_{\theta,j}(\mathbf h_t),\\
\widehat v_j^{\mathrm{current}}(t)&=v_j^{\mathrm{meas}}(t),
\quad\text{if $j$ is observed},\\
\widehat v_j^{\mathrm{current}}(t)
&=a_{\theta,j}(\mathbf v_{\rightarrow j}^{\mathrm{input}}(t))\\
&\quad+n_{\theta,j}\bigl(
\mathbf v_{\rightarrow j}^{\mathrm{input}}(t),
\widehat v_j^{-}(t),\mathbf h_t\bigr),
\quad\text{otherwise}.
\end{aligned}
\label{eq:architecture}
\end{equation}
We chose an LSTM for processing windows of a fixed \emph{history length}. Its hidden state $\mathbf h_t$ and memory state $\mathbf c_t$ summarize the past complete vectors, which contain measurements and reconstructions. Output head $d_{\theta,j}$ turns this summary into the one-step prediction of variable $j$. An unobserved variable is the sum of an affine estimate $a_{\theta,j}(\mathbf x)=\mathbf w_j^\top\mathbf x+\gamma_j$ and a nonlinear correction by the one-hidden-layer network $n_{\theta,j}$. All trainable weights form $\theta$.
 
The \emph{learned nominal trajectory} starts $\delta t$ samples before the current time $t$ and predicts the entire interval up to $t$ using the learned normal model, without corrections from process measurements. This open-loop reuses the same LSTM and heads:
\begin{equation}
\begin{aligned}
\widehat{\mathbf v}^{\mathrm{nominal}}(t\mid t-\delta t)
&=D_{\theta}^{(\delta t)}\bigl(
\widehat{\mathbf v}^{\mathrm{current}}(t-\delta t),\\
&\hspace{13mm}\mathbf h_{t-\delta t},\mathbf c_{t-\delta t},
\mathbf u_{t-\delta t+1:t}\bigr).
\end{aligned}
\label{eq:observers}
\end{equation}
Operator $D_{\theta}^{(\delta t)}$ applies the model $\delta t$ times to its own outputs. Commands or setpoints $\mathbf u_{t-\delta t+1:t}$ may enter if they do not reveal the fault. The start at $t-\delta t$ can already be fault-affected, so the nominal trajectory is no guaranteed fault-free reference. The \emph{diagnosis horizon} $\delta t$ can differ from the history length and the training horizon.
 
The network is trained on normal data with
\begin{equation}
\mathcal L_{\mathrm{base}}=
\lambda_{\mathrm{current}}\mathcal L_{\mathrm{current}}+
\lambda_{\mathrm{nominal}}\mathcal L_{\mathrm{nominal}}.
\label{eq:base-loss}
\end{equation}
$\mathcal L_{\mathrm{current}}$ compares reconstructions with offline targets and is dropped if none exist. $\mathcal L_{\mathrm{nominal}}$ compares the rollout with normal data. In \emph{recursive training}, the model feeds its predictions back as inputs and learns from errors across several steps. The loss covers every step of a fixed \emph{training horizon}, and gradients pass through all steps. One-step training is a faster approximation that uses only the next prediction. Diagnosis always uses the full rollout over $\delta t$. The weights $\lambda_{\mathrm{current}},\lambda_{\mathrm{nominal}}\geq0$ balance both terms.  
After training, the weights are frozen. 
 
\subsection{Using A Priori Knowledge to Improve Virtual Observability}
\label{sec:godel-training}
 
\begin{table*}[t]
\centering
\normalsize
\caption{A priori knowledge for training, with process-industry examples. Experiments use the applicable rule groups.}
\label{tab:knowledge}
\setlength{\tabcolsep}{4.5pt}
\begin{tabular}{@{}p{0.23\textwidth}p{0.72\textwidth}@{}}
\toprule
Prior knowledge & Example relation \\
\midrule
Topology & Connected variables have compatible normal reconstruction errors. \\
Qualitative relations & Variables change in the same or opposite directions under specified conditions. \\
Temporal relations & Responses follow within a specified delay. \\
Physical flow & Signed flows balance and mass flow agrees with volume flow through density. \\
Modes and state dynamics & Modes are exclusive. Filling implies a nondecreasing tank level. \\
Sensor physics & Measurement equations relate states and outputs, such as level and pressure. \\
\bottomrule
\end{tabular}
\end{table*}

Several behavior models may fit normal data equally well yet disagree outside the observed operating range. This ambiguity is known as \emph{underspecification} \citep{damour2022underspecification}. For a virtual pump-speed sensor, matching measured flows alone may not determine the physical speed uniquely. Supplied physical and system knowledge adds an \emph{inductive bias}: training favors reconstructions consistent with known plant relationships. Appropriate knowledge can reduce the training data required and improve extrapolation to operating conditions not covered by those data \citep{oneto2025informed}. These benefits depend on the rules and are not guaranteed.

Table~\ref{tab:knowledge} lists the main types of knowledge. The knowledge is compiled as rules over \emph{predicates}, that is, elementary relations between process variables such as ``the tank level rises''. During training, the selected rules $\mathcal K_{\mathrm{train}}$ are evaluated on both trajectories.
 G\"odel logic assigns each predicate $p$ a truth degree between zero and one. Conjunction takes the minimum, disjunction the maximum, and implication follows G\"odel semantics:
\begin{equation}
\begin{aligned}
\tau(p)&=\left(1+\rho_p/\sigma_p\right)^{-1},\\
\tau(p\Rightarrow_G q)&=
\begin{cases}
1,&\tau(p)\leq\tau(q),\\
\tau(q),&\text{otherwise.}
\end{cases}
\end{aligned}
\label{eq:goedel}
\end{equation}
Here, $\rho_p\geq0$ measures the violation of $p$, in physical units for measurement equations, and $\sigma_p>0$ is its engineering tolerance.
 
Each rule is evaluated separately at each sample of a training sequence. These rule evaluations enter training as a penalty:
\begin{equation}
\mathcal L=\mathcal L_{\mathrm{base}}+
\beta\sum_{g\in\mathcal G}\bar\alpha_g\,
\frac{1}{|\mathcal K_g|}
\sum_{k\in\mathcal K_g}w_k\,\overline\ell_k.
\label{eq:loss}
\end{equation}
For rule $k$, $\overline\ell_k$ is the mean shortfall $\max\{0,m-\tau(k)\}$ below a required truth degree $m\in(0,1]$ over all usable rule evaluations, including satisfied ones. Set $\mathcal G$ contains the rule groups with usable rules, and $\mathcal K_g\subseteq\mathcal K_{\mathrm{train}}$ the rules of group $g$. Averaging within each group keeps groups with many rules from dominating. The weights $\beta\geq0$, $\bar\alpha_g$, and $w_k>0$ are fixed in advance. With $\beta=0$, Eq.~\eqref{eq:loss} reduces to Eq.~\eqref{eq:base-loss}.
 
A rule evaluation is \emph{usable} if the required history and operating conditions are available at that sample. Other evaluations are left out rather than counted as satisfied.
 
\subsection{Time-Synchronous Virtual-Observability Diagnosis}
\label{sec:diagnosis}
 
ReDiVO compares local symptom-onset times to identify the root cause. After training and calibration, Algorithm~\ref{alg:redivo} applies three diagnosis steps at scheduled \emph{evaluation times}.
 
\textbf{1. Detect anomalies by comparing value ranges} (Algorithm~\ref{alg:redivo}, line~\ref{line:anomaly}). To account for noise and estimation errors, each measured or reconstructed process value and its nominal prediction receive uncertainty ranges:
\begin{equation}
\begin{aligned}
C_j(t)&=\bigl[\widehat v_j^{\mathrm{current}}(t)\pm E_j^{\mathrm C}\bigr],\\
N_j(t)&=\bigl[\widehat v_j^{\mathrm{nominal}}(t\mid t-\delta t)\pm E_j^{\mathrm N}\bigr],
\end{aligned}
\label{eq:value-ranges}
\end{equation}
with $[x\pm E]=[x-E,x+E]$. Bound $E_j^{\mathrm C}\geq0$ covers measurement or reconstruction error, and $E_j^{\mathrm N}\geq0$ covers prediction error over $\delta t$. Overlapping ranges give no evidence of a fault. Separated ranges give the normalized gap
\begin{equation}
U_j(t)=\frac{1}{b_j}\max\!\left\{0,
\underline C_j(t)-\overline N_j(t),
\underline N_j(t)-\overline C_j(t)\right\},
\label{eq:residual}
\end{equation}
where underlines and overlines denote lower and upper limits. The scale $b_j>0$ makes gaps comparable across variables.
 
\textbf{2. Confirm persistent symptoms.} Single exceedances can be outliers. Variable $j\in B_r$ therefore shows a symptom of candidate $r$ only if $U_j(t)$ exceeds the threshold $\eta_{r,j}>0$ at $P$ consecutive evaluation times (Algorithm~\ref{alg:redivo}, line~\ref{line:symptoms}). The timestamp of the first one is the onset time $\widehat t_{r,j}$, and the $P$th confirms the symptom. A candidate with at least one confirmed symptom is \emph{symptomatic}.
 
\textbf{3. Compare symptom-onset times} (Algorithm~\ref{alg:redivo}, lines~\ref{line:onset}--\ref{line:decision}). Each onset is evaluated on the range $[\widehat t_{r,j}\pm\zeta_{r,j}]$, with $\zeta_{r,j}\geq0$ in physical time. The candidate interval $\mathcal I_r=[t_r^-,t_r^+]$ takes the smallest lower and the smallest upper limit over the confirmed symptoms in $B_r$. A non-symptomatic candidate $q$ receives $t_q^-=+\infty$ for the comparison only, which is no evidence of health. Candidate $r$ is the unique diagnosis if it is certainly earlier than every competitor:
\begin{equation}
t_r^+<t_q^- \qquad
\text{for every }q\in\mathcal R\setminus\{r\}.
\label{eq:onset-separation}
\end{equation}
Otherwise, if symptoms exist, the result is an \emph{ambiguity group}. It contains every symptomatic candidate whose lower limit does not exceed the smallest upper limit, including touching intervals. For a unique diagnosis, the same rule yields only $r$. The \emph{Potentially correct} metric in Section~\ref{sec:eval} uses this group. Without symptoms, ReDiVO returns no diagnosis.
 
\begin{algorithm}[t]
\caption{ReDiVO training and diagnosis}
\label{alg:redivo}
\small
\begin{algorithmic}[1]
\Require Normal training, validation, and calibration data, predictors, rules $\mathcal K_{\mathrm{train}}$, root-cause variables $B_r$, horizons $\mathcal D$, persistence $P$
\Ensure Unique candidate, ambiguity group, or no diagnosis
\Statex \textit{Training on normal data}
\State Train $\theta$ with Eq.~\eqref{eq:loss} and select the epoch
\State Freeze $\theta$, select $\delta t\in\mathcal D$, calibrate bounds and thresholds
\Statex \textit{Diagnosis of one episode}
\For{each evaluation time $t$}
  \State Compute both trajectories and anomaly gaps $U_j(t)$ with Eqs.~\eqref{eq:architecture}--\eqref{eq:observers}, \eqref{eq:value-ranges}--\eqref{eq:residual} (Step~1)\label{line:anomaly}
\EndFor
\State Confirm persistent symptoms and record onset times (Step~2)\label{line:symptoms}
\State Form candidate onset-time intervals $\mathcal I_r$ (Step~3)\label{line:onset}
\State \Return $r$ satisfying Eq.~\eqref{eq:onset-separation}, else the ambiguity group, or no diagnosis without symptomatic candidates (Step~3)\label{line:decision}
\end{algorithmic}
\end{algorithm}

\section{Evaluation}
\label{sec:eval}

Virtual sensor reconstruction, the effect of prior knowledge, and onset-based fault isolation are evaluated on artificial data and three diagnosis datasets: the Extended Tennessee Eastman Process (TEP) \citep{bathelt2015revision,reinartz2021extended}, HAI-CPPS v2.1, and the TwinFlow dataset~\cite{hranisavljevic2026twinflow,hranisavljevic2026dataset}.

Under the selection procedure of Section~\ref{sec:observer}, reconstruction loss on normal validation data selects the training epoch, with prediction loss resolving ties. After freezing the weights, the diagnosis horizon minimizes the mean absolute current--nominal difference at the forecast endpoint over the predefined horizons. Separate normal recordings calibrate error bounds and thresholds. All choices are fixed before fault testing. The HAI study uses the modified procedure described in Section~\ref{sec:hai-v21-evaluation}.

\emph{Correct} counts true-cause inclusion in the earliest persistent symptom-time group, including ties.
\emph{Potentially correct} counts use Section~3.4's possible-cause group. 
\emph{Uniquely correct} counts require only the true cause in that group. 
Hit@3 counts inclusion among the first three active candidates. 
Cases without symptoms count as failures.
Alphabetical ties convey no fault-type evidence. 
Random references select from all candidates while matching the method's output sizes, including empty outputs.

In the result tables, \emph{detection} counts fault recordings with at least one persistent diagnostic symptom, whereas \emph{healthy alarms} counts fault-free recordings with such a symptom. Detection alone does not establish the right cause or an alarm after the fault begins. Reconstruction errors use the targets and units stated in each caption.

\subsection{Artificial Data}
\label{sec:synthetic-evaluation}

Two artificial studies test reconstruction, masking, and prior knowledge, without claiming industrial realism. Their results remain separate.

\subsubsection{Linear Component Networks}
\label{sec:linear-evaluation}

Each component has two hidden states, six redundant outputs, linear discrete-time dynamics, and delayed connections. With four outputs hidden, the two remaining independent measurements still constrain both states. Eight graph seeds and 24 conditions yield 912 models, with separate normal data for training, selection, and calibration.

\begin{table}[t]
\centering
\normalsize
\caption{Linear networks, eight seeds. Setup: component count or combined stress at 30 components. Mask: hidden outputs per component. Full: value, direction, and delay rules. Rates in percent over 192 healthy and 240 fault recordings per condition (960 faults at 120 components).}
\label{tab:linear-results}
\setlength{\tabcolsep}{2.5pt}
\begin{tabular}{@{}llrrrr@{}}
\toprule
Setup & Rules & Mask & \shortstack{Unique\\diagnosis} & Detection & \shortstack{Healthy\\alarms} \\
\midrule
30 & None & 2 & 99.17 & 100.00 & 13.02 \\
30 & Full & 2 & 100.00 & 100.00 & 11.46 \\
\midrule
30 & None & 4 & 99.58 & 100.00 & 9.38 \\
30 & Full & 4 & 100.00 & 100.00 & 11.46 \\
\midrule
120 & None & 2 & 96.77 & 100.00 & 43.75 \\
120 & Full & 2 & 97.71 & 100.00 & 38.54 \\
\midrule
Stress & None & 4 & 47.08 & 55.00 & 7.81 \\
Stress & Value & 4 & 65.83 & 72.50 & 6.25 \\
Stress & Full & 4 & 59.58 & 65.00 & 7.81 \\
\bottomrule
\end{tabular}
\end{table}

Isolation remains high under masking and scaling (Table~\ref{tab:linear-results}), but healthy alarms increase with system size. Affine reconstruction is more accurate on these linear data.

Under combined masking, noise, delays, coupling, and weak faults, value rules improve isolation from 47.08\% to 65.83\%. The 95\% gain interval is 13.75--23.33 percentage points from 20,000 paired seed-bootstrap resamples, but the Holm-adjusted probability is 0.703 across 90 comparisons. Only 1.60\% of faulty time points have confirmed symptoms, distinguishing recording-level diagnosis from continuous detection.

\subsubsection{Random Hybrid Automata}
\label{sec:hybrid-evaluation}

Eight randomly parameterized six-component systems switch between three modes with affine or nonlinear ordinary differential equations (ODEs). Both internal states remain hidden, with two or four of six measurements masked. Actuator-effectiveness loss and additional outflow occur as steps or ramps near mode switches or in locally stationary intervals.

Per system, 24 normal recordings train the models, eight select them, and 64 calibrate them. Tests use 16 healthy recordings and 48 fault cases, each lasting 200 seconds at 0.2-second sampling. Two model initializations are averaged within each independent system.

\begin{table}[t]
\centering
\normalsize
\caption{Hybrid automata. Rates in percent. Mask: hidden measurements per component. Both states are reconstructed. Rules affect training only.}
\label{tab:hybrid-results}
\setlength{\tabcolsep}{3pt}
\begin{tabular}{@{}lrrrr@{}}
\toprule
Rules & Mask & \shortstack{Unique\\diagnosis} & Detection & \shortstack{Healthy\\alarms} \\
\midrule
None & 2 & 72.40 & 83.20 & 11.72 \\
Structure & 2 & 70.96 & 82.16 & 13.67 \\
Mode & 2 & 52.73 & 61.33 & 12.50 \\
Both & 2 & 56.51 & 64.97 & 10.94 \\
\midrule
None & 4 & 64.97 & 82.29 & 13.28 \\
Structure & 4 & 62.24 & 80.99 & 10.55 \\
Mode & 4 & 68.36 & 84.77 & 18.36 \\
Both & 4 & 63.54 & 79.69 & 12.89 \\
\bottomrule
\end{tabular}
\end{table}

Structure rules align reconstruction errors across connections. Mode rules encode mode-dependent measurement equations and mode exclusivity. Each group retains weight 0.5 when used alone or together. Without rules, virtual sensors support correct isolation in 72.40\% and 64.97\% of cases under the two masks (Table~\ref{tab:hybrid-results}).

On healthy test recordings with four measurements masked, combined rules reduce current-state mean absolute error (MAE) from 0.003910 to 0.003345, but increase nominal-forecast MAE from 0.032515 to 0.045318, in normalized ODE-state units. The diagnosis change is $-1.43$ percentage points, with a descriptive 95\% interval of $[-8.72,8.33]$ from 5,000 paired system-bootstrap resamples stratified by dynamics family. Structure alone does not improve reconstruction. Mode-dependent ridge reconstruction is more accurate, but no ridge diagnosis pipeline was tested.

\subsection{Extended TEP}
\label{sec:tep-evaluation}

Extended TEP \citep{bathelt2015revision,reinartz2021extended} provides 412 fault cases from four modes. The 24 runs combine four rule settings, two masks, and three seeds. 
Windows span 44 samples. 

Table~\ref{tab:tep-results} excludes 32 cases post hoc: condenser-valve sticking (IDV15) and stripper steam-transfer variation (IDV16), whose sensors never activate persistently for unresolved reasons. 
All 26 candidates remain. 
Including those cases gives 59.71\% Correct, 74.92\% Potential, and 64.40\% Hit@3 without rules or masking.

\begin{table*}[t]
\centering
\normalsize
\caption{TEP means (\%), three model seeds and 380 fault cases. Full combines topology and qualitative rules. Mask counts additionally hidden channels. Correct allows ties in the earliest symptom group and is not unique isolation. Detection counts a persistent symptom anywhere in a fault window. Healthy-recording alarm rates were not evaluated (n.e.). Random matches the no-rule output sizes and has no detector.}
\label{tab:tep-results}
\setlength{\tabcolsep}{6pt}
\begin{tabular}{@{}lrrrrrr@{}}
\toprule
G\"odel rules & Mask & Correct & Potential & Hit@3 & Detection & \shortstack{Healthy\\alarms} \\
\midrule
None & 0 & 64.74 & 81.23 & 69.82 & 93.86 & n.e. \\
Full & 0 & 64.39 & 79.65 & 67.72 & 93.68 & n.e. \\
Random & 0 & 7.86 & 30.79 & 10.20 & -- & -- \\
\midrule
None & 3 & 66.84 & 80.88 & 69.65 & 93.77 & n.e. \\
Full & 3 & 66.32 & 79.82 & 68.86 & 93.33 & n.e. \\
Random & 3 & 8.65 & 30.94 & 10.30 & -- & -- \\
\bottomrule
\end{tabular}
\end{table*}

\begin{table*}[t]
\centering
\small
\setlength{\tabcolsep}{3.5pt}
\renewcommand{\arraystretch}{1.12}
\caption{Results for ReDiVO on the HAI-CPPS v2.1 benchmark.
All rows use G\"odel regularization ($\beta=0.01$), state specific calibration, and separate regression heads.
Healthy alarms count one healthy episode. Reconstruction and forecast MAE use normal-validation pump speeds in pump rpm, weighting all pumps and operating phases equally.}
\label{tab:hai-final-combined}

\begin{tabular}{llrrrrr}
\hline
Dataset & Training & \begin{tabular}[c]{@{}r@{}}Correct\\ pump\end{tabular} & Detection & \begin{tabular}[c]{@{}r@{}}Healthy\\ alarms\end{tabular} & \begin{tabular}[c]{@{}r@{}}Reconstruction\\ MAE\end{tabular} & \begin{tabular}[c]{@{}r@{}}Forecast\\ MAE\end{tabular} \\ \hline
ds5 & Recursive & 4/4 & 4/4 & 0/1 & 7.644 & 9.745 \\
ds5 & One-step & 4/4 & 4/4 & 0/1 & 9.314 & 12.728 \\
ds6 & Recursive & 6/6 & 6/6 & 0/1 & 6.160 & 9.859 \\
ds6 & One-step & 6/6 & 6/6 & 0/1 & 9.811 & 16.307 \\
ds7 & Recursive & 3/4 & 3/4 & 0/1 & 7.440 & 10.784 \\
ds7 & One-step & 2/4 & 4/4 & 1/1 & 8.543 & 22.046 \\
ds8 & Recursive & 6/6 & 6/6 & 1/1 & 6.597 & 10.412 \\
ds8 & One-step & 5/6 & 5/6 & 0/1 & 9.155 & 18.008 \\
ds9 & Recursive & 8/8 & 8/8 & 1/1 & 5.459 & 9.653 \\
ds9 & One-step & 6/8 & 8/8 & 1/1 & 8.650 & 18.033 \\
ds10 & Recursive & 2/2 & 2/2 & 0/1 & 5.725 & 10.178 \\
ds10 & One-step & 2/2 & 2/2 & 0/1 & 7.502 & 13.374 \\ \hline
\textbf{All} & \textbf{Recursive} & \textbf{29/30} & \textbf{29/30} & \textbf{2/6} & \textbf{6.287} & \textbf{10.066} \\
\textbf{All} & \textbf{One-step} & \textbf{25/30} & \textbf{29/30} & \textbf{2/6} & \textbf{8.735} & \textbf{16.562} \\ \hline
\end{tabular}

\end{table*}

Masking hides recycle flow, stripper underflow, and condenser cooling flow. 
Earliest groups average 2.04--2.25 candidates without rules, possible groups 8.01--8.05. 
Inclusion does not imply unique isolation.

Full rules change unmasked Correct by $-0.35$ percentage points, with a 95\% interval of $[-1.58,0.96]$ from 2,000 paired case-bootstrap resamples without multiplicity correction. Six qualitative runs failed gradient verification, and their replay remained incomplete. Tests use known fault windows, not continuous monitoring.

\subsection{HAI-CPPS Version 2.1}
\label{sec:hai-v21-evaluation}

The HAI-CPPS dataset is a simulation of a modular industrial process plant.
We focus on datasets ds5 to ds10, using all 30 supplied faulty recordings and one healthy test recording per dataset.
Specifically, we focus on the pumps within each module, as they provide a good evaluation testbed for ReDiVO's virtual sensor reconstruction. 
The model estimates effective pump speeds from other measurements, without observing these speeds directly.

We evaluate ReDiVO with G\"odel regularization during training, a state dependent calibration, and additional separate regression heads.
The masked pump speed from the dataset provides the offline targets and validation references.

\textbf{Diagnosis.} Recursive training identifies the correct pump in 29/30 faulty recordings. 
All 29 decisions identify a unique pump, the remaining recording produces no diagnosis. 
One-step training identifies the correct pump in 25/30 cases.
Both versions raise a pump alarm in 29/30 faulty recordings
and give false alarms in two of the six healthy recordings (Table~\ref{tab:hai-final-combined}).

\textbf{Prognosis and reconstruction.} Recursive training predicts healthy validation behavior more accurately.
The forecast MAE falls from 16.562 to 10.066~rpm (39.2\%), and reconstruction MAE from 8.735 to 6.287 rpm for the pump speed.

\subsection{TwinFlow}
\label{sec:twinflow-evaluation}

TwinFlow is a material-handling simulation with conveyors, robots, mobile transport vehicles, and workstations. 
Faults include motor failures, emergency stops, and conveyor speed-calibration errors \citep{hranisavljevic2026twinflow}.
Here, we focus on conveyor speed-calibration faults, which affect the accuracy of the reported conveyor speed.

Across all 15 speed-calibration faults in the TwinFlow dataset, ReDiVO identifies the correct faulty conveyor
in 12 cases (80\%), while considering the full set of 1,444 possible
component--fault combinations. 
This means that the diagnosis narrows the problem down to the right conveyor, although several explanations for its abnormal behavior may remain.

Finding the faulty component and identifying its exact fault type are different tasks. 
ReDiVO determines both uniquely in four of the 15 events, as TwinFlow's fault types do overlap in the characterization of symptoms. 
E.g. a speed calibration fault has the same symptoms as a motor fault. 

\subsection{Ablation Studies: Temporal Resolution and Fault Types}
\label{sec:temporal-fault-ablations}

Each artificial ablation uses six independent graph seeds.

\emph{Temporal resolution and clock offsets.} Sampling every four rather than one second retained every strong fault in the earliest group without rules. Correction learned from normal synchronization pulses restored ordering disrupted by clock offsets. This tests symptom timestamps, not asynchronous observer inputs or short transients.

\emph{Fault types.} Strong additive faults were reliably localized. Noise and masking impaired weaker gain and decay faults. Duplicated fault labels remained ambiguous. Sensor-identity, sum, difference, and direction rules improved some reconstructions but suppressed weak symptoms at weight one. Better reconstruction did not ensure better diagnosis.

False alarms limit continuous monitoring. Paired differences are not significant after Holm correction.

\subsection{Discussion of RQ1--RQ3}
\label{sec:rq-discussion}

\textbf{RQ1: Can virtual sensors provide useful local symptoms?} Yes, under the tested observation conditions. Both artificial studies reconstruct hidden states even when four of six process measurements are unavailable. Isolation remains nearly complete in the linear tests, and the reconstructed states enable useful diagnosis in the hybrid tests. In HAI, virtual pump-speed sensors identify the faulty pump in 29 of 30 fault recordings with recursive training, compared with 25 of 30 with one-step training. Recursive training also improves normal-validation reconstruction and forecasting. Thus, root-cause variables need not all be measured directly when the remaining measurements sufficiently constrain them.

\textbf{RQ2: Which prior knowledge helps?} Helpful rules depend on the relationships and observation conditions. Adding more rules is not automatically beneficial. Value equalities and known sign reversals reduce reconstruction error in the linear tests and improve diagnosis in the difficult condition, although the diagnosis gain is not significant after campaign-wide correction. Direction and delay rules provide no comparable diagnosis gain in that condition.

In the hybrid tests, mode-dependent measurement rules improve current-state reconstruction, whereas structure compatibility alone does not. Combined rules improve reconstruction under stronger masking but worsen nominal forecasting, without a reliable diagnosis gain. With two measurements masked, combined rules leave more faults without a diagnosis than training without rules. Lower reconstruction error alone is therefore insufficient for choosing the rules. TEP likewise shows no consistent diagnosis gain: qualitative rules were nearly inactive, while active topology rules did not improve isolation. HAI uses G\"odel rules in both training variants, so this comparison does not isolate their benefit.

\textbf{RQ3: When does the earliest symptom identify the cause?} The method provides useful isolation when the local fault produces a persistent symptom before competing symptoms. Without rules, it uniquely diagnoses 72.40\% and 64.97\% of fault recordings in the six-component hybrid tests under the two masks. Additional-outflow faults are isolated more reliably than actuator-effectiveness losses, which more often fail to activate a persistent local symptom. The failure audit also links early false symptoms to nominal-forecast errors around normal mode switches, even when current-state reconstruction is accurate. Synchronization alone cannot remove these prediction and detection errors.

The linear tests retain high isolation under masking and moderate disturbances, and clock-offset correction restores symptom ordering in the temporal ablation. In HAI, recursive training uniquely identifies the faulty pump in 29 of 30 fault recordings. TEP includes the true cause in the earliest symptom group in about two thirds of cases, substantially above size-matched random selection, but ties can retain several causes. The remaining priorities are accurate nominal forecasts, reliable local symptom detection, and temporal separation between competing symptoms.

\section{Conclusion}
\label{sec:conclusion}

ReDiVO connects reconstruction of named root-cause variables with time-based fault isolation. Measurements and virtual sensors form the current system trajectory. Comparison with a learned nominal trajectory provides persistent local symptoms. Their onset times distinguish a unique cause or a group of possible causes. G\"odel regularization incorporates physical and system knowledge during training only.

\textbf{RQ1} is supported by useful virtual-sensor symptoms in both artificial studies and the HAI pump cases, including four hidden measurements out of six in the artificial systems. Direct measurement of every root-cause variable is not necessary under these observation conditions. \textbf{RQ2} has a conditional answer. Specific value relationships reduce reconstruction error in the reported linear comparisons. Mode-dependent rules improve reconstruction in the hybrid systems, but combining them with structure rules can worsen nominal forecasting and diagnosis. TEP shows no consistent additional diagnosis gain. Priors must therefore be assessed for their effect on both trajectories and the resulting symptoms. For \textbf{RQ3}, onset-based diagnosis achieves high isolation in the linear tests, correctly isolates most faults in the six-component hybrid tests without rules, and selects the faulty pump in 29 of 30 HAI fault recordings with recursive training, compared with 25 of 30 with one-step training. TEP places the true cause in the earliest symptom group substantially more often than random selection. Missed actuator symptoms and early false symptoms remain important limits of the hybrid results, despite useful reconstruction.

Future work should improve nominal forecasts around mode switches and calibrate false alarms using longer normal recordings. Accumulating small residual deviations may reveal faults missed by consecutive-threshold detection, but requires normal-data calibration and checks for shifted symptom-onset times. Further ablations should identify which G\"odel rules help which variables and fault mechanisms, and at what weights. Timing studies should vary clock errors, sampling intervals, and sensor-response delays to quantify the precision needed to distinguish competing symptom onsets. Independent fault realizations should test whether these findings generalize.

\medskip
\noindent\textbf{Code and data availability.}
Algorithm code and generators for the artificial datasets are provided at
\url{https://github.com/ollini42/research-artifacts/tree/d989de6e4737cf836c54fa2846b5e023cff2aace/goedel-diagnosis-2026}.
Extended TEP data are available at \url{https://doi.org/10.11583/DTU.13385936.v1}.
HAI-CPPS benchmark data \citep{ehrhardt2025hai} are available at \url{https://doi.org/10.21227/5ewb-cn40},
with v2.1 generation code at \url{https://github.com/j-ehrhardt/hai-cpps-benchmark/tree/v2.1}.
Twinflow data are available at \url{https://zenodo.org/records/22943653}, 
with the generation code at \url{https://github.com/imb-hsu/twinflow}.

\section*{Disclaimer}
During the preparation of this work the authors used OpenAI's ChatGPT and
Codex in order to translate texts, discuss concepts and fine-tune the
readability of the manuscript and to support the development and review of
research code. After using this tool/service, the authors reviewed and edited
the content as needed and take full responsibility for the content of the
published article.

\bibliographystyle{ifacconf}
\bibliography{paper_references}

@article{damour2022underspecification,
  author = {D'Amour, Alexander and Heller, Katherine and Moldovan, Dan and others},
  title = {Underspecification Presents Challenges for Credibility in Modern Machine Learning},
  journal = {Journal of Machine Learning Research},
  volume = {23},
  number = {226},
  pages = {1--61},
  year = {2022},
  url = {https://jmlr.org/papers/v23/20-1335.html},
}

@article{oneto2025informed,
  author = {Oneto, Luca and Ridella, Sandro and Anguita, Davide},
  title = {Informed Machine Learning: Excess Risk and Generalization},
  journal = {Neurocomputing},
  volume = {646},
  pages = {130521},
  year = {2025},
  doi = {10.1016/j.neucom.2025.130521},
}

@article{isermann2005model,
  author =        {Isermann, Rolf},
  journal =       {Annual Reviews in Control},
  number =        {1},
  pages =         {71--85},
  title =         {Model-Based Fault-Detection and Diagnosis: Status and
                   Applications},
  volume =        {29},
  year =          {2005},
  doi =           {10.1016/j.arcontrol.2004.12.002},
}

@misc{ehrhardt2025hai,
  author = {Ehrhardt, Jonas and Moddemann, Lukas and Diedrich, Alexander and Niggemann, Oliver},
  title = {{HAI-CPPS}: The {Hamburg} {AI} Benchmark for Cyber-Physical Production Sytems},
  year = {2025},
  howpublished = {IEEE Dataport},
  doi = {10.21227/5ewb-cn40},
}

@article{krysander2008efficient,
  author =        {Krysander, Mattias and {\AA}slund, Jan and
                   Nyberg, Mattias},
  journal =       {IEEE Trans. Systems, Man, and Cybernetics -- Part A},
  number =        {1},
  pages =         {197--206},
  title =         {An Efficient Algorithm for Finding Minimal
                   Overconstrained Subsystems for Model-Based Diagnosis},
  volume =        {38},
  year =          {2008},
}

@article{qin2012survey,
  author =        {Qin, S. Joe},
  journal =       {Annual Reviews in Control},
  number =        {2},
  pages =         {220--234},
  title =         {Survey on Data-Driven Industrial Process Monitoring
                   and Diagnosis},
  volume =        {36},
  year =          {2012},
}

@article{revach2022kalmannet,
  author =        {Revach, Guy and Shlezinger, Nir and Ni, Xiaoyong and
                   L{\'o}pez Escoriza, Adri{\`a} and
                   van Sloun, Ruud J. G. and Eldar, Yonina C.},
  journal =       {IEEE Transactions on Signal Processing},
  pages =         {1532--1547},
  title =         {{KalmanNet}: Neural Network Aided Kalman Filtering
                   for Partially Known Dynamics},
  volume =        {70},
  year =          {2022},
  doi =           {10.1109/TSP.2022.3158588},
}

@inproceedings{ludwig2025virtualsensors,
  author =        {Ludwig, Bjoern and Ehrhardt, Jonas and
                   Niggemann, Oliver},
  booktitle =     {Proc. IEEE 30th International Conference on Emerging
                   Technologies and Factory Automation},
  pages =         {1--8},
  title =         {Creating Virtual Sensors Using Neural Networks},
  year =          {2025},
  doi =           {10.1109/ETFA65518.2025.11205672},
}

@inproceedings{assaad2023easyrca,
  author =        {Assaad, Charles K. and Ez-Zejjari, Imad and Zan, Lei},
  booktitle =     {Proceedings of the 26th International Conference on
                   Artificial Intelligence and Statistics},
  pages =         {8395--8404},
  publisher =     {PMLR},
  series =        {Proceedings of Machine Learning Research},
  title =         {Root Cause Identification for Collective Anomalies in
                   Time Series Given an Acyclic Summary Causal Graph
                   with Loops},
  volume =        {206},
  year =          {2023},
  url =           {https://proceedings.mlr.press/v206/assaad23a.html},
}

@inproceedings{han2025aerca,
  author =        {Han, Xiao and Absar, Saima and Zhang, Lu and
                   Yuan, Shuhan},
  booktitle =     {The Thirteenth International Conference on Learning
                   Representations},
  title =         {Root Cause Analysis of Anomalies in Multivariate Time
                   Series through {Granger} Causal Discovery},
  year =          {2025},
  url =           {https://openreview.net/forum?id=k38Th3x4d9},
}

@article{badreddine2022logic,
  author =        {Badreddine, Samy and d'Avila Garcez, Artur and
                   Serafini, Luciano and Spranger, Michael},
  journal =       {Artificial Intelligence},
  pages =         {103649},
  title =         {Logic Tensor Networks},
  volume =        {303},
  year =          {2022},
  doi =           {10.1016/j.artint.2021.103649},
}

@article{vanKrieken2022analyzing,
  author =        {van Krieken, Emile and Acar, Erman and
                   van Harmelen, Frank},
  journal =       {Artificial Intelligence},
  pages =         {103602},
  title =         {Analyzing Differentiable Fuzzy Logic Operators},
  volume =        {302},
  year =          {2022},
}

@article{hogea2024logiclstm,
  author =        {Hogea, Eduard and Onchi\c{s}, Darian M. and
                   Yan, Ruqiang and Zhou, Zheng},
  journal =       {Journal of Manufacturing Systems},
  pages =         {892--902},
  title =         {{LogicLSTM}: Logically-Driven Long Short-Term Memory
                   Model for Fault Diagnosis in Gearboxes},
  volume =        {77},
  year =          {2024},
  doi =           {10.1016/j.jmsy.2024.10.003},
}

@article{reinartz2021extended,
  author =        {Reinartz, Christopher Clarc and Kulahci, Murat and
                   Ravn, Ole},
  journal =       {Computers \& Chemical Engineering},
  pages =         {107281},
  title =         {An Extended Tennessee Eastman Simulation Dataset for
                   Fault-Detection and Decision Support Systems},
  volume =        {149},
  year =          {2021},
  doi =           {10.1016/j.compchemeng.2021.107281},
}

@article{bathelt2015revision,
  author =        {Bathelt, Andreas and Ricker, N. Lawrence and
                   Jelali, Mohieddine},
  journal =       {IFAC-PapersOnLine},
  number =        {8},
  pages =         {309--314},
  title =         {Revision of the Tennessee Eastman Process Model},
  volume =        {48},
  year =          {2015},
  doi =           {10.1016/j.ifacol.2015.08.199},
}

@article{sampath1995diagnosability,
  author =        {Sampath, Meera and Sengupta, Raja and
                   Lafortune, St{\'e}phane and Sinnamohideen, Kasim and
                   Teneketzis, Demosthenis},
  journal =       {IEEE Transactions on Automatic Control},
  number =        {9},
  pages =         {1555--1575},
  title =         {Diagnosability of Discrete-Event Systems},
  volume =        {40},
  year =          {1995},
  doi =           {10.1109/9.412626},
}

@article{harirchi2018guaranteed,
  author =        {Harirchi, Farshad and Ozay, Necmiye},
  journal =       {Automatica},
  pages =         {476--488},
  title =         {Guaranteed Model-Based Fault Detection in
                   Cyber-Physical Systems: A Model Invalidation Approach},
  volume =        {93},
  year =          {2018},
  doi =           {10.1016/j.automatica.2018.03.040},
}

@article{diedrich2024learning,
  author  = {Diedrich, Alexander and Moddemann, Lukas and Niggemann, Oliver},
  title   = {Learning system descriptions for cyber-physical systems},
  journal = {IFAC-PapersOnLine},
  volume  = {58},
  number  = {4},
  pages   = {628--633},
  year    = {2024},
  doi     = {10.1016/j.ifacol.2024.07.289},
}

@article{steude2024diagnosis,
  author  = {Steude, Henrik Sebastian and Moddemann, Lukas and Diedrich, Alexander and Ehrhardt, Jonas and Niggemann, Oliver},
  title   = {Diagnosis driven anomaly detection for cyber-physical systems},
  journal = {IFAC-PapersOnLine},
  volume  = {58},
  number  = {4},
  pages   = {13--18},
  year    = {2024},
  doi     = {10.1016/j.ifacol.2024.07.186},
}

@inproceedings{diedrich2024sensor,
  author    = {Diedrich, Alexander and Heesch, Ren{\'e} and Bozzano, Marco and Ludwig, Bj{\"o}rn and Cimatti, Alessandro and Niggemann, Oliver},
  title     = {Inferring Sensor Placement Using Critical Pairs and Satisfiability Modulo Theory},
  booktitle = {35th International Conference on Principles of Diagnosis and Resilient Systems (DX 2024)},
  series    = {OASIcs},
  volume    = {125},
  pages     = {9:1--9:19},
  year      = {2024},
  doi       = {10.4230/OASIcs.DX.2024.9},
}

@inproceedings{moddemann2024extracting,
  author    = {Moddemann, Lukas and Steude, Henrik Sebastian and Diedrich, Alexander and Pill, Ingo and Niggemann, Oliver},
  title     = {Extracting Knowledge using Machine Learning for Anomaly Detection and Root-Cause Diagnosis},
  booktitle = {2024 IEEE 29th International Conference on Emerging Technologies and Factory Automation (ETFA)},
  publisher = {IEEE},
  year      = {2024},
  doi       = {10.1109/ETFA61755.2024.10710647},
}

@InProceedings{hranisavljevic2026twinflow,
  author = {Hranisavljevic, Nemanja and Diedrich, Alexander and Moddemann, Lukas and
            Schaeffer, Domenic and Marek, Frank and Pill, Ingo and Niggemann, Oliver},
  title = {{HSU TwinFlow: A Living Benchmark for Evaluating Anomaly Detection and
            Diagnosis Methods in Cyber-Physical Production Systems Using
            Digital-Twin-Generated Data}},
  booktitle = {37th International Conference on Principles of Diagnosis and
               Resilient Systems (DX 2026)},
  pages = {7:1--7:14},
  ISBN = {978-3-95977-455-0},
  ISSN = {2190-6807},
  year = {2026},
  volume = {148}
}

@misc{hranisavljevic2026dataset,
  doi = {10.5281/ZENODO.22943652},
  url = {https://zenodo.org/doi/10.5281/zenodo.22943652},
  author = {Hranisavljevic,  Nemanja and Diedrich,  Alexander and Moddemann,  Lukas and Schaeffer,  Domenic and Marek,  Frank and Pill,  Ingo and Niggemann,  Oliver},
  language = {en},
  title = {HSU TwinFlow: A Living Benchmark for Evaluating Anomaly Detection and Diagnosis in CPPSs Using Digital-Twin-Generated Data},
  publisher = {Zenodo},
  year = {2026},
  copyright = {Creative Commons Attribution 4.0 International}
}

\end{document}